# AI-Driven Clinical Decision Support System for Enhanced Diabetes Diagnosis and Management

Mujeeb Ur Rehman *Senior Member, IEEE*, Imran Rehan, Sohail Khalid, *Senior Member, IEEE*

*Abstract*—Identifying type 2 diabetes mellitus can be challenging, particularly for primary care physicians. Clinical decision support systems incorporating artificial intelligence (AI-CDSS) can assist medical professionals in diagnosing type 2 diabetes with high accuracy. This study aimed to assess an AI-CDSS specifically developed for the diagnosis of type 2 diabetes by employing a hybrid approach that integrates expert-driven insights with machine learning techniques. The AI-CDSS was developed (training dataset: n = 650) and tested (test dataset: n = 648) using a dataset of 1298 patients with and without type 2 diabetes. To generate predictions, the algorithm utilized key features such as body mass index, plasma fasting glucose, and hemoglobin A1C. Furthermore, a clinical pilot study involving 105 patients was conducted to assess the diagnostic accuracy of the system in comparison to non-endocrinology specialists. The AI-CDSS showed a high degree of accuracy, with 99.8% accuracy in predicting diabetes, 99.3% in predicting prediabetes, 99.2% in identifying at-risk individuals, and 98.8% in predicting no diabetes. The test dataset revealed a 98.8% agreement between endocrinology specialists and the AI-CDSS. Type 2 diabetes was identified in 45% of 105 individuals in the pilot study. Compared with diabetes specialists, the AI-CDSS scored a 98.5% concordance rate, greatly exceeding that of non-endocrinology specialists, who had an 85% agreement rate. These findings indicate that the AI-CDSS has the potential to be a useful tool for accurately identifying type 2 diabetes, especially in situations in which diabetes specialists are not readily available.

## I. Introduction

The study of medical diagnoses is essential in the healthcare sector to enhance the clinical outcomes. A disease is a condition or set of conditions that results in pain and suffering, illness, malfunction, or ultimately, death [1]. The pathological process, or systematic study of a disease, involves identifying diseases based on clinical experts' interpretations of signs and symptoms [2]. Assessing the pathology of a disease requires a diagnosis [3], which is the process of

Mujeeb Ur Rehman is with the School of Computer Science and Informatics, De Montfort University, Leicester, UK. (e-mail: mujeeb.rehman@dmu.ac.uk).

Imran rehan is with the department of physics, Islamia College Peshawar, Pakistan. (e-mail: Irehanyousfzai@gmail.com).

Sohail Khalid is with the Computer Science Department, University of Management and Technology, Lahore, Pakistan (e-mail: sohail.khalid@umt.edu.pk).

determining its identity based on its clinical presentation, as illustrated in Fig. 1.

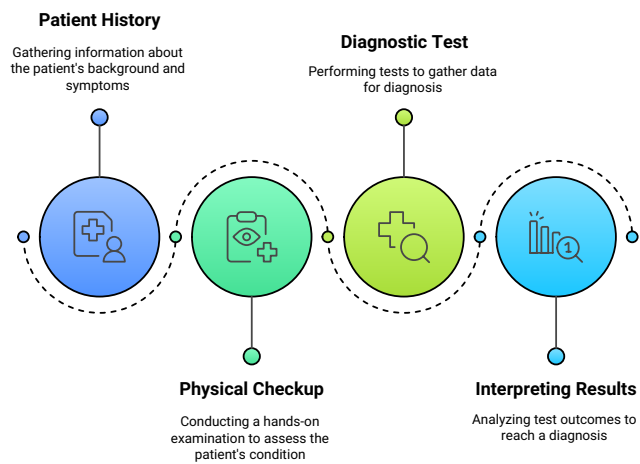

Fig. 1: Block diagram showing the steps involved in diagnosis.

Diabetes is a serious chronic illness that affects glucose metabolism, and can have major consequences if left untreated. Over the last few decades, it has become a chronic medical problem worldwide. Insulin regulates blood glucose levels, and diabetes is mostly caused by the insufficient and inefficient use of this hormone. Diabetes can cause several health problems that affect numerous body parts. In 2013, approximately 382 million individuals worldwide were reported to have diabetes. It is anticipated that the number of individuals with diabetes will sharply increase to 592 million by 2035 [4]. Diabetes affects most people in developing countries [5]. Many factors, including dietary habits, environmental conditions, inflammatory processes, genetics, and lifestyle choices, can lead to diabetes. These intricate interactions between the components influence the mechanisms and development of diabetes are influenced by these intricate interactions between components [6]. Although there are theories linking oxidative damage, obesity, and aging to the onset of diabetes, the precise underlying causes remain unknown. The World Health Organization (WHO) has classified diabetes mellitus (DM) as a serious concern owing to its sharp increase in prevalence. There are three primary forms of diabetes: type 1, type 2, and gestational diabetes. Type 1 diabetes develops when the body's immune system mistakenly attacks the pancreatic beta cells and disrupts insulin production. Gestational diabetes develops during pregnancy and typically disappears after delivery [7], [8], [9]. This may increase the risk of problems for both mothers and children. Of the three varieties of diabetes, type 2 diabetes is the most prevalent and is invariably associated with lifestyle factors. When a person has type 2 diabetes, their body develops insulin resistance because the pancreas is unable to generate enough insulin to fulfil their needs. Consequently, the body can no longer regulate blood glucose levels, and to keep on track, people with type 2 diabetes also need to take other drugs and continue their current practices [10], [11]. The symptoms of diabetes vary widely, depending on the type of diabetes. However, frequent urination, unexplained weight loss, increased hunger, exhaustion, blurred vision, numbness, and repeated infections are the typical symptoms. Increased thirst and urination are common early signs of diabetes; additionally, they may lose weight even when their appetite is high [12]. High blood sugar levels affect the eye lens and cause blurred vision. It can also weaken the immune system, increase the risk of infection, and complicate the healing of wounds. Various methods exist for controlling blood glucose levels depending on the type of diabetes. Insulin therapy is typically used to control type 1 diabetes. However, oral medication or insulin injections may be necessary for type 2 diabetes depending on its severity [13], [14], [15]. Patients with diabetes can manage their blood glucose levels by frequently checking them and altering their prescription schedules, as needed. Diet was also important in this case. Artificial intelligence (AI) has significantly influenced the growth of several sectors, particularly in the medical field, by integrating machine-learning techniques [16], [17]. There are two types of AI decision algorithms in the market: white boxes and black boxes. A white-box AI-driven decision system is primarily built using supervised algorithms such as the decision tree algorithm, and incorporates associations and visibility among rules for the analysis of accumulated data. In contrast, black-box AI algorithms are opaque, and it is challenging to understand the logic and process used to arrive at the corresponding conclusions. IBM Watson for Oncology (WFO) is an example of a black-box AI-based decision support system. For the recommended course of treatment for breast cancer, WFO showed a 93 % concordance rate; however, it was unable to provide the recommended methods used to reach the ultimate clinical conclusion. A medical information system called the Clinical Decision Support System (CDSS) helps doctors make clinical decisions. Six decades ago, the concept of computer-based medical choices was established in informatics [18]. Although considerable enthusiasm exists for using AI to advance the CDSS, the practical challenges of clinical implementation hinder their rapid development. An effective CDSS must match each patient's distinctive features to the clinical database, offer patient-centric assessments and suggestions, and then convey the recommendations in a white-box format so that the physicians may make the ultimate decision. [19], [20] The purpose of this study was to evaluate the degree of agreement between diabetes specialists and the AI-CDSS in diagnosing type 2 diabetes and identifying individuals with diabetes, prediabetes, and normal glucose levels. Initially, a hybrid technique combining machine-learning (ML)-driven rule generation and expert-driven knowledge acquisition was used to construct an AI-CDSS. Second, as a pilot clinical trial, the diagnostic correlation (level of concordance) of the AI-CDSS was assessed in a test group of individuals with and without type 2 diabetes. Finally, a prospective evaluation of AI-CDSS's diagnostic efficacy of AI-CDSS was conducted in a series of patients who visited the outpatient clinic with diabetes-related symptoms.

## II. AI-CDSS Algorithms for Diabetes

The AI-CDSS for diabetes diagnosis combines data-driven machine learning insights with rule-based expert knowledge. Table I lists the steps adopted in the AI-CDSS methodology.

### A. Collecting and preprocessing data

Demographic, clinical, and laboratory factors were included in data collected from a typical group of patients. Data from this dataset included client age, sex, body mass index (BMI), HbA1c values, fasting plasma glucose levels, and other pertinent clinical parameters. The data were preprocessed and cleaned, missing values were handled, and feature normalization was performed. The normalization was performed using

$$x' = \frac{X - \mu}{\sigma} \quad (1)$$

where $X$ represents the feature value, $\mu$ is the mean of the feature, and $\sigma$ is the standard deviation of the feature. The missing values were handled using the imputation method as

$$x_{imputed} = \frac{\sum_{i=1}^{n} x_i}{n} \quad (2)$$

where $x_i$ represents the available values and $n$ represents the number of available values.

### B. Expert-Driven Knowledge Acquisition

Endocrinology experts were consulted to determine the most important diagnostic standard for type 2 diabetes. These standards were converted into organized forms such as decision trees and mind maps. The decision tree was finalized as the Diabetes Clinical Knowledge Model (D-CKM) after iterative expert feedback. For example, a decision tree can be mathematically described as follows:

$$D(x) = \begin{cases} 1 & \text{if } x \in \text{ leaf node with } P(\text{Diabetes } = 1) > 0.5 \\ 0 & \text{otherwise} \end{cases} \quad (3)$$

### C. Rule Generation Driven by Machine Learning

The dataset was subjected to several machine learning methods, including Decision Tree (DT), Random Forest, J48, Chi-squared Automatic Interaction Detection (CHAID), and Classification and Regression Tree (CART). The most important features for diabetes diagnosis were found using feature selection methods such as Recursive Feature Elimination (RFE):

$$RFE(X, y, k) = \underset{\text{subset of features}}{arg\ min} \text{ ModelError }(X_{\text{subset}}, y) \quad (4)$$

TABLE I: Steps Involved in Proposed CDSS Method

| Pseudo Algorithm of CDSS Method |
|---|
| **Step-I: Data Collection and Preprocessing** |
| FUNCTION |
| preprocess_data(patient_data) |
| Handle missing values |
| Normalize features |
| RETURN preprocessed_data |
| END FUNCTION |
| **Step-II: Expert-Driven Knowledge Acquisition** |
| FUNCTION |
| create_expert_model() |
| Consult endocrinology experts |
| Define diagnostic criteria and contributing factors |
| Construct decision tree (D-CKM) |
| RETURN D-CKM |
| END FUNCTION |
| **Step-III: Machine Learning-Driven Rule Generation** |
| FUNCTION |
| create_ml_model(training_data) |
| Apply machine learning algorithms |
| Perform feature selection |
| Evaluate and rank algorithms |
| Select top-performing algorithm |
| Train the selected model |
| RETURN trained_model |
| ENDFUNCTION |
| **Step-IV: Hybridization and Integration** |
| FUNCTION |
| create_hybrid_model(D-CKM, trained_model) |
| Combine expert-driven and ML-driven models |
| Integrate decision pathways and rules |
| RETURN hybrid_model (R-CKM) |
| ENDFUNCTION |
| **Step-V: Clinical Validation** |
| FUNCTION |
| validate_model(test_data, hybrid_model) |
| Apply hybrid model to test data |
| Compare with expert diagnoses |
| Calculate performance metrics |
| Perform subgroup analysis |
| RETURN performance_results |
| ENDFUNCTION |
| **Step-VI: Main Program** |
| BEGIN |
| Collect and preprocess patient data patient_data = READ_DATA() |
| preprocessed_data =preprocess_data(patient_data) |
| Split data into training and testing sets training_data |
| test_data = SPLIT_DATA(preprocessed_data) |
| Create expert-driven & ML-driven models D-CKM = create_expert_model() |
| trained_model = create_ML_model(training_data) |
| Create hybrid model hybrid_model = create_hybrid_model (D-CKM, trained_model) |
| Validate the hybrid model performance_results = validate_model(test_data, hybrid_model) |
| Display or report the results PRINT performance_results |
| END |

where k denotes the number of features chosen, *y* denotes the target variable, and *X* denotes the feature set. Based on the accuracy of the algorithms, number of rules, and other characteristics, the algorithms were assessed and ranked.

$$\text{Rank} = \frac{\text{Accuracy} + \frac{1}{\text{Number of Rules}} + \frac{1}{\text{Number of Attributes}}}{3} \quad (5)$$

The top-performing algorithm, CART, was selected to develop the ML-driven guidance system.

### D. Validation and Training of Models

Stratified sampling was used to divide the dataset into training (70 %) and testing (30 %) subsets while preserving class distribution. The training set of data was used to train the algorithm.

$$\theta = \arg\min_{\theta} Loss\left(h_\theta\left(X_{\text{train}}\right), y_{\text{train}}\right) \quad (6)$$

where $h_\theta$ represents the hypothesis and $\theta$ the model parameters. The test data were used to validate the model, and the performance measures (specificity, sensitivity, and accuracy) were assessed [21].

$$\text{Accuracy} = \frac{TP + TN}{TP + TN + FP + FN} \quad (7)$$

$$\text{Sensitivity} = \frac{TP}{TP + FN} \quad (8)$$

$$\text{Specificity} = \frac{TN}{TN + FP} \quad (9)$$

where the numbers of true positives, true negatives, false positives, and false negatives are denoted by TP, TN, FP, and FN, respectively.

### E. Hybridization and Integration

A hybrid AI-CDSS was developed by combining ML-driven- and expert-driven decision trees. The system was effectively integrated using both rule- and data-driven insights. The following is a mathematical representation of this integration:

$$H(x) = \alpha \cdot D_{CKM}(x) + (1 - \alpha) \cdot h_\theta(x) \quad (10)$$

where $H(x)$ represents the hybrid model, $h_\theta(x)$, and the expert-driven model's weight by $\alpha$.

### F. Clinical Assessment and Pilot Research

A prospective clinical pilot study was conducted to evaluate AI-CDSS in an actual environment. The diagnostic accuracy of the AI-CDSS was compared with that of physicians who were not specialists. Cohen's kappa coefficient was used to evaluate the degree of agreement among the AI-CDSS and expert diagnoses:

$$\kappa = \frac{p_o - p_e}{1 - p_e} \quad (11)$$

where $p_e$ is the predicted agreement by chance and $p_o$ is the measured agreement.

## III. RESULTS

### A. AI-CDSS development for diabetes

The AI-CDSS was developed using predetermined phases, including expert-driven information collection, ML-driven procedure formation, and the hybridization of both forms of knowledge, employing training datasets of 650 individuals with and without diabetes.

## B. Expert-driven information gathering

The clinical suggestions for detecting type 2 diabetes were initially converted into mind maps and subsequently into decision trees during the knowledge modelling phase. Four outcomes were possible: verified diabetes, prediabetes, at risk, and no diabetes. The final decision tree, referred to as the R-CKM, contained 14 contributing elements (see Table II).

The flow shown in Algorithm 1 starts with the patient's first symptoms and family history, proceeds through several clinical parameters, including blood pressure, BMI, HbA1c, and fasting glucose, and ends with specific suggestions.

**Algorithm 1** Clinical Symptoms Classification

1: **Input:** Patient Data
2: **Output:** Classified Clinical Symptoms
3: **if** Family History = Yes **then**
4:      Mark as Positive History
5: **else**
6:      Mark as Negative History
7: **end if**
8: **if** Physical Activity = High **then**
9:      Mark as Active
10: **else**
11:      Mark as Inactive
12: **end if**
13: **if** $HbA1c \geq 6.5\%$ **then**
14:      Mark as Elevated HbA1c
15: **else**
16:      Normal HbA1c
17: **end if**
18: **if** $FastingGlucose \geq 126$ mg/dL **then**
19:      Mark as High Glucose
20: **else**
21:      Normal Glucose
22: **end if**
23: **if** $BMI > 25$ **then**
24:      Mark as Overweight
25: **else**
26:      Normal Weight
27: **end if**
28: **if** Blood Pressure = Hypertension **then**
29:      Mark as High BP
30: **else**
31:      Normal BP
32: **end if**
33: **Output:** Classified Clinical Symptoms

Symptomatology Results in: History of the Family "Yes" or "No" Physical Activity "High" / "Low", HbA1c ">6.5%" / "≤ 6.5%", Fasting Blood Sugar "> 126 mg/dL" / "≤ 126 mg/dL", BMI **"> 25" / **"≤ 25" Blood Pressure "Hypertension" versus "Normal".

TABLE II: List of Diabetes-Related Contributing Factors to CKM

| SN | Name of Attribute | Description of Attributes |
|---|---|---|
| 1 | Symptoms & Signs | The patient presents with symptoms including weariness, increased thirst, and frequent urination. |
| 2 | Medical Background | Comprises the patient's medical history, including any history of diabetes in the family and hypertension, while applying steroids or other drugs. |
| 3 | Physical Inspection and examination results, including acanthosis nigricans, blood pressure, waist circumference, and BMI. | |
| 4 | Fasting Plasma Glucose | Blood tests to determine glucose levels following a minimum of eight hours of fasting. |
| 5 | Oral Glucose Tolerance Test | Testing the blood for glucose levels both before and after ingesting a sweet drink. |
| 6 | Hemoglobin A1c | Test for blood that measures the average blood sugar levels during the previous three months. |
| 7 | Lipid Profile | Blood tests that measure triglyceride, HDL, and LDL cholesterol levels. |
| 8 | Tests for kidney function and creatinine, eGFR, and the ratio of albumin to creatinine in urine. | |
| 9 | Insulin Levels | Test for insulin resistance in the blood that measures insulin levels during fasting. |
| 10 | C-Peptide | C-peptide levels in the blood is measured to distinguish between type 1 and type 2 diabetes. |
| 11 | Beta-cell Function | Evaluation of pancreatic beta-cell activity using HOMA-B and other assays. |
| 12 | Insulin Resistance | Evaluation through the use of tests, for example, Homeostatic Model Assessment of Insulin Resistance. |
| 13 | Autoantibodies | Autoantibody tests, such as ZnT8, IA-2, and GAD65, to identify type 1 diabetes. |
| 14 | Family Background | Includes details about immediate family members with diabetes or other related diseases. |

## C. ML-based rule development

CHAID, DT, J48, and CART are the five machine-learning techniques that we used. These algorithms identified important characteristics such as fasting plasma glucose, BMI, and HbA1c levels as major contributing factors. Using an auto-feature selection technique, the most important features were age, BMI, fasting plasma glucose, and HbA1c to improve the model performance, as depicted in Fig. 2.

Based on the accuracy, the number of extracted rules, and the number of attributes involved, we ranked each algorithm. Owing to its maximum accuracy (89.8%) and ranking value (0.798), the CART algorithm was selected to produce ML-driven knowledge. The main features of the CART algorithm were BMI and fasting plasma glucose and HbA1c levels. With an accuracy of 99.8 %, it predicted diabetes in individuals, 99.3 % predicted prediabetes, 99.2 % predicted at-risk, and 98.8 % predicted no diabetes. The contributing factors are presented in Table II.

The levels of accuracy displayed by the five algorithms vary, as shown in Table III. The DT visualization of the CART is presented in Algorithm 2.

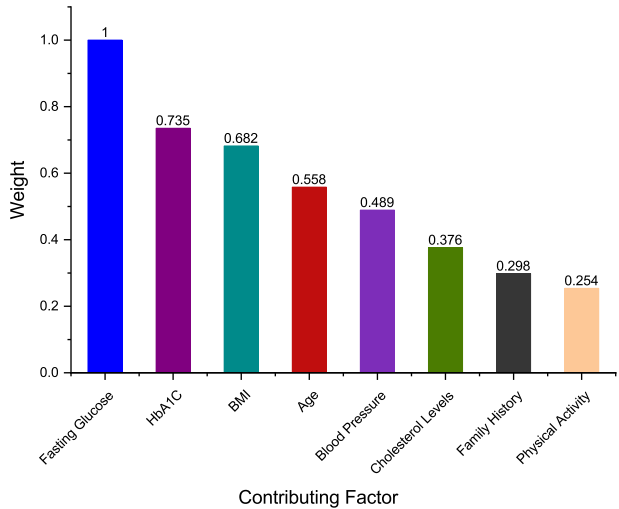

Fig. 2: Machine learning-selected contributing elements in the ML-driven method.

---

**Algorithm 2** Diabetes Classification Based on HbA1c, Glucose, and BMI
---
1: **Input:** HbA1c, Fasting Glucose, BMI
2: **Output:** Diabetes Classification with Accuracy
3: **if** $HbA1c \leq 6.5\%$ **then**
4:     No Diabetes (Accuracy: 98.8%)
5: **else**
6:     **if** $FastingGlucose > 126$ mg/dL **then**
7:         Individuals at Risk (Accuracy: 99.2%)
8:         **if** $BMI > 25$ **then**
9:             Type 2 Diabetes (Accuracy: 99.8%)
10:         **else**
11:             Prediabetes (Accuracy: 99.3%)
12:         **end if**
13:     **else**
14:         No Diabetes (Accuracy: 98.8%)
15:     **end if**
16: **end if**
17: **Output:** Classification with Accuracy

---

### D. Hybrid knowledge

The final hybrid knowledge, the Refined-Clinical Knowledge Model (R-CKM) for diabetic disease, was produced by combining the Clinical Knowledge Model (CKM) from the expert-driven knowledge strategy with the Prediction Model (PM) from the ML-driven knowledge approach. This procedure guarantees a thorough examination of the patients, resulting in more precise recommendations. During CKM construction, doctors may overlook certain qualities or paths of attributes. In these cases, the ML-generated PM identifies the absent attributes or paths. For example, CKM may prioritize Clinical Symptoms (as shown in Algorithm 3), whereas PM may begin by assessing HbA1c and fasting glucose readings. By integrating these pathways, the hybridization algorithm expands the number of rules and enhances the coverage of the patient cases. Between the Clinical Symptoms and HbA1c attributes, the hybridization algorithm detects that CKM lacks a path of unavailable values. The number of established rules in CKM increased significantly with the addition of this new approach. By incorporating additional pathways, R-CKM covers more patient instances, improves the accuracy, and produces more precise recommendations.

TABLE III: Characteristics chosen as contributing elements by various ML algorithms.

| Machine Learning Algorithms | Number of attributes | Structure of Features |
|---|---|---|
| CART | 3 | A) Fasting Plasma Glucose<br>B) HbA1c<br>C) BMI |
| J48 | 6 | A) HbA1c<br>B) Fasting Plasma Glucose<br>C) Blood Pressure<br>D) Family History<br>E) BMI<br>F) Age |
| Random Forest | 5 | A) HbA1c<br>B) BMI<br>C) Fasting Plasma Glucose<br>D) Cholesterol Levels<br>E) Age |
| Decision Tree | 4 | A) BMI,<br>B) HbA1c,<br>C) Physical Activity,<br>D) Fasting Plasma Glucose |
| CHAID | 4 | A) HbA1c<br>B) BMI<br>C) Fasting Plasma Glucose<br>D) Age |

## IV. AI-CDSS FOR DIABETES VALIDATION

### A. Analyzed population

There were 648 patients in the test dataset (531 with type 2 diabetes and 117 without disease). Individuals diagnosed with type 2 diabetes were of greater age (55.2 ± 14.2 years compared to 45.3 ± 12.6, P < 0.001), had a greater percentage of males (55 vs. 41%, P=0.010), and had a higher probability of having a family history of diabetes (68.5% versus 20.4%, P<0.001). Individuals diagnosed with type 2 diabetes also had higher BMI (30.2 ± 3.2 kg/$m^2$ versus 24.5 ± 3.2 kg/$m^2$, P <0.001), raised blood pressure (140/90 ± 15/10 mmHg versus 120/80 ± 10/5 mmHg), and higher HbA1c levels (7.5 ± 1.2% versus 5.2 ± 0.4%, P < 0.001). Diabetes patients had substantially higher fasting plasma glucose levels (150.2 ± 35.61 mg/dL) than those with 95.4 ± 10.3 mg/dL (P < 0.001). Furthermore, those diagnosed with type 2 diabetes exhibited reduced HDL cholesterol levels (40.3 ± 10.5 mg/dL in contrast to 55.2 ± 12.1 mg/dL, P < 0.001) and elevated triglyceride levels (200.10 ± 60.40 mg/dL in contrast to 100.5 ± 20.30 mg/dL, P < 0.001). They also reported lower levels of adherence to a nutritious diet (30% compared to 65%, P < 0.001) and physical activity (36% compared to 76%, P < 0.001). Furthermore, individuals with diabetes were more likely to use drugs to treat hypertension (50% vs. 16%, P < 0.001). (See Table IV) The symptoms of polyuria and polydipsia were more common among people with type 2 diabetes (45.7% vs. 5.2%, P < 0.001), and their waist circumference was larger (105.5 ± 10.5 cm vs. 85.4 ± 7.3 cm, P < 0.001).

**Algorithm 3** Clinical Symptoms Decision Flow
1: **Input:** Patient Data
2: **Output:** Classified Clinical Symptoms
3: **if** Family History = Yes **then**
4:     Mark as Positive History
5: **else if** Family History = No **then**
6:     Mark as Negative History
7: **end if**
8: **if** Physical Activity = High **then**
9:     Mark as Active
10: **else if** Physical Activity = Low **then**
11:     Mark as Inactive
12: **end if**
13: **if** $HbA1c \geq 6.5\%$ **then**
14:     Mark as Elevated HbA1c
15: **else**
16:     Normal HbA1c
17: **end if**
18: **if** $FastingGlucose \geq 126$ mg/dL **then**
19:     Mark as High Glucose
20: **else**
21:     Normal Glucose
22: **end if**
23: **if** $BMI > 25$ **then**
24:     Mark as Overweight
25: **else**
26:     Normal Weight
27: **end if**
28: **if** Blood Pressure = Hypertension **then**
29:     Mark as High BP
30: **else**
31:     Normal BP
32: **end if**
33: **Output:** Classified Clinical Symptoms

### B. Accuracy of diagnostics

The comparative analysis results are presented in Fig. 3. The expert-driven AI-CDSS, ML-driven AI-CDSS, and hybrid CDSS (Proposed) provide concordance rates for diabetes classification for verified diabetes, with concordance rates of 95 %, 96 %, and 99.8 % for the Expert-Driven AI-CDSS, ML-Driven AI-CDSS, and Hybrid CDSS models, respectively. Similarly, for Prediabetes, Expert-Driven AI-CDSS yields a concordance rate of 92 %, ML-Driven AI-CDSS yields 93 %, and Hybrid CDSS 99.3 % respectively (see Table V – VII). Further, for the At-Risk group, the Expert-Driven AI-CDSS provided a concordance of 90 %, whereas the ML-driven AI-CDSS provided 91 % and for Hybrid CDSS is 99.2 %, respectively. Additionally, for individuals without diabetes, the expert-driven AI-CDSS has a rate of 88 %, the ML-driven AI-CDSS has a rate of 89 %, and the Hybrid CDSS has a rate of 98.8 %. For all classifications, the Expert-Driven AI-CDSS resulted in 91 %, where ML-Driven AI CDSS outcomes as 92 %, and Hybrid CDSS resulted in 99 % (see Table V – VII).

TABLE IV: Table showing the characteristics of the study inhabitants (n = 648).

| Features | Not having Diabetes | Having Diabetes type 2 | P* value |
|---|---|---|---|
| Age in years | 45.4 ± 12.6 | 55.2 ± 14.3 | <0.001 |
| Male percentage (%) | 41 | 55 | 0.010 |
| Diabetes in Family History (%) | 20.4 | 68.5 | <0.001 |
| Body Mass Index (kg/m2) | 24.5 ± 3.2 | 30.2 ± 4.8 | <0.001 |
| Plasma glucose during fasting (mg/dL) | 95.4 ± 10.3 | 150.2 + 35.61 | <0.001 |
| The percentage of HbA1c | 5.2 ± 0.4 | 7.5 ± 1.2 | <0.001 |
| Blood Pressure (mmHg) | 120/80 + 10/5 | 140/90 + 15/10 | <0.001 |
| Triglycerides (mg/dL) | 100.5 ± 20.30 | 200.10 ± 60.40 | <0.001 |
| The mg/dL of triglycerides | 55.2 ± 12.1 | 40.3 ± 10.5 | <0.001 |
| Activity Level in Physical Terms (%) | 76 | 36 | <0.001 |
| Dietary Practices (% Balanced Diet) | 65 | 30 | <0.001 |
| Hypertension Medication (%) | 16 | 50 | <0.001 |
| Circumference of the Waist (cm) | 85.4 ± 7.3 | 105.5 ± 10.5 | <0.001 |
| Signs (such as polyuria and polydipsia) (%) | 5.2 | 45.7 | <0.001 |

*P value represents the statistical significance between the groups.

TABLE V: Table showing confusing matrix for Expert-Driven Artificial Intelligence CDSS

| Observations | Verified Diabetics | Prediabetes | At-Risk | Non Diabetics | Correct (%) | Sensitivity | Specificity |
|---|---|---|---|---|---|---|---|
| Verified Diabetics | 195 | 0 | 0 | 0 | 95 | 0.96 | 1 |
| Pre-diabetics | 0 | 148 | 13 | 20 | 92 | 0.86 | 0.92 |
| At-Risk | 0 | 13 | 117 | 13 | 90 | 0.87 | 0.88 |
| No Diabetics | 0 | 0 | 13 | 89 | 88 | 0.71 | 0.95 |
| Percentiles | 95 % | 92 % | 90 % | 88 % | 91 | 0.71 | 0.96 |

Furthermore, 98.8% of the endocrinology specialists in the test dataset agreed with the Hybrid CDSS, demonstrating the system's high reliability. The AI-CDSS considerably outperformed non-endocrinology specialists, who had an 85% agreement rate and a 98.5% confidence rate in a clinical pilot study of 105 patients, 45 of whom were diagnosed with type 2 diabetes.

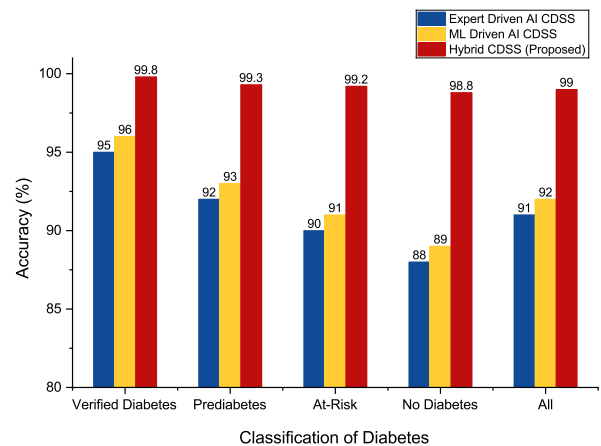

Fig. 3: Comparative evaluation of various methods' diagnostic accuracy in the retrospective cohort.

These findings demonstrate how well the Hybrid CDSS can diagnose type 2 diabetes, making it a useful tool in clinical contexts where access to specialized knowledge may be constrained (see Tables V – VII).

TABLE VI: Table shows a confusing matrix for ML-driven Artificial Intelligence CDSS.

| Observations | Verified Diabetics | Prediabetes | At-Risk | Non Diabetics | Correct (%) | Sensitivity | Specificity |
|---|---|---|---|---|---|---|---|
| Verified Diabetics | 195 | 0 | 0 | 0 | 96 | 0.96 | 1 |
| Pre-diabetics | 0 | 151 | 12 | 0 | 93 | 0.87 | 0.93 |
| At-Risk | 0 | 10 | 118 | 11 | 91 | 0.90 | 0.94 |
| No Diabetics | 0 | 0 | 22 | 87 | 89 | 0.72 | 0.94 |
| Percentiles | 96 % | 93 % | 91 % | 89 % | 92 | 0.72 | 0.94 |

TABLE VII: The table shows a confusing matrix for Hybrid CDSS (Proposed) for Diabetes Disease.

| Observations | Verified Diabetics | Prediabetes | At-Risk | Non Diabetics | Correct (%) | Sensitivity | Specificity |
|---|---|---|---|---|---|---|---|
| Verified Diabetics | 212 | 0 | 0 | 0 | 99.8 | 0.97 | 1 |
| Pre-diabetics | 0 | 154 | 0 | 3 | 99.3 | 0.90 | 0.97 |
| At-Risk | 0 | 0 | 120 | 1 | 99.2 | 0.94 | 0.99 |
| No Diabetics | 0 | 0 | 4 | 108 | 98.8 | 0.92 | 0.97 |
| Percentiles | 99.8-% | 99.3-% | 99.2-% | 98.8-% | 99.0 | 0.94 | 0.99 |

The sensitivity and specificity shown by the expert-driven, ML-driven, and hybrid approaches were as follows: with expert guidance, the AI-CDSS produced results with a sensitivity of 0.96 and a specificity of 0.95. The specificity and sensitivity of the ML-driven AI-CDSS were both 0.94. The suggested hybrid CDSS, which integrates machine-learning and expert-driven methodologies, produced results with a 0.99 specificity and a sensitivity of 0.94. To facilitate clarity, Fig. 4 presents an illustration of sensitivity and specificity among several CDSS approaches.

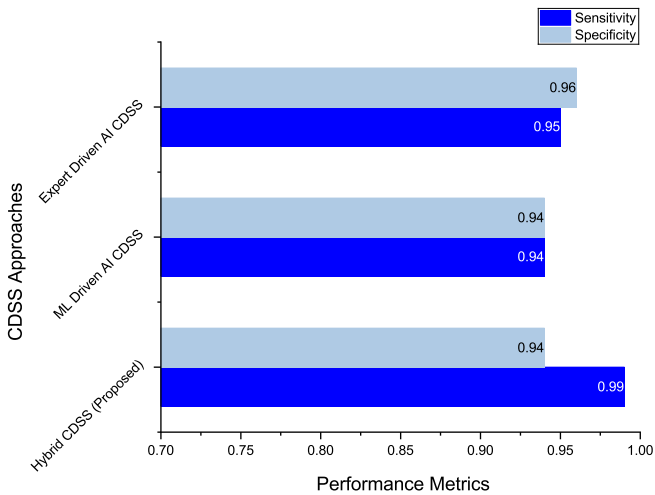

Fig. 4: Sensitivity and Specificity Comparison of Various CDSS Methodologies

The sensitivity and specificity of three distinct Clinical Decision Support Systems (CDSS) approach; Expert-Driven AI-CDSS, ML-Driven AI-CDSS, and the Hybrid CDSS (Proposed) are depicted in this picture. Compared with other approaches, the Hybrid CDSS performs better at reliably detecting illnesses because it combines machine learning with expert-driven methodologies. It exhibited the highest specificity (0.99) and highest sensitivity (0.94). A comparison

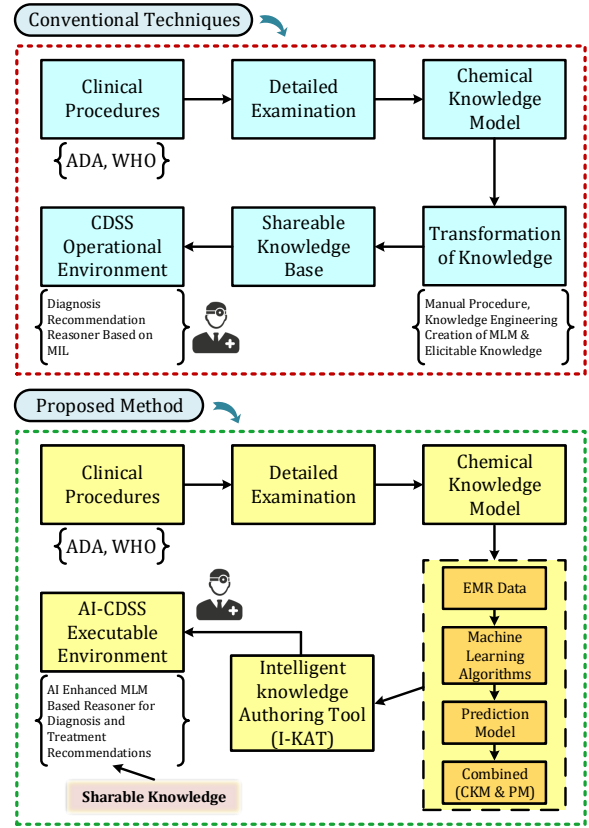

Fig. 5: Figure showing an association between the current CDSSs and our AI-based CDSS method.

between the current CDSSs and our AI-based CDSS approach is shown in Fig. 5.

### C. Sub-group investigation

We categorized the patients into smaller groups according to the important diagnostic criteria to assess the effectiveness of the AI-CDSS in various scenarios. Set A comprised all pertinent clinical and laboratory data, whereas Set B concentrated on key diagnostic indicators such as BMI, HbA1c, and fasting plasma glucose. A thorough examination of the data is essential for precise diagnosis, as demonstrated by the constant decrease in the concordance rate between Set B and Set A (See Fig. 6).

The individuals involved in our study were between the ages of 20 and 100. The accuracy of the system remained high, with no fluctuations across age groups, suggesting that age had no discernible impact on the functionality of the system (Table VIII).

TABLE VIII: Table breaking down the diagnostic accuracy by age group of patients.

| Age | Diabetic individuals | Classified Correctly | Classified False |
|---|---|---|---|
| 20-40 years | 25 | 24 | 1 |
| 41-60 years | 85 | 84 | 1 |
| 61-80 years | 320 | 315 | 5 |
| 81-100 years | 180 | 176 | 4 |

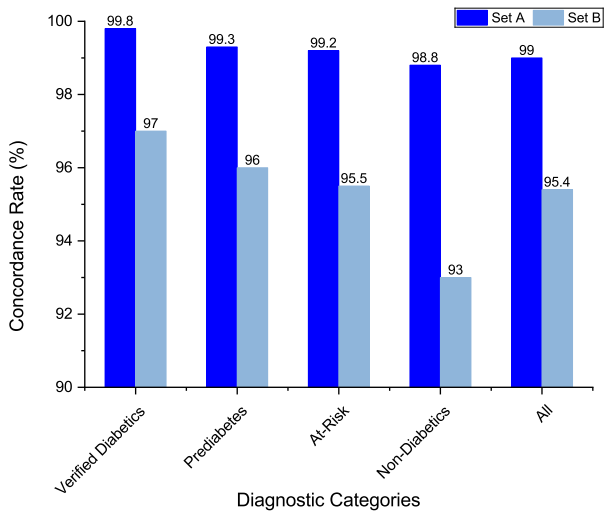

Fig. 6: Figure showing a comparison of the rate of concordance of AI–CDSS between Set A and Set B.

This study included 105 participants who visited the outpatient clinic in a row and had symptoms that could indicate diabetes. A total of 102 patients were included in the final study after eliminating three partial data points from three patients. Among these, type 2 diabetes was identified in 45%. The concordance rate of non-endocrinology specialists in this prospective population was 85%, while the AI-CDSS attained a concordance rate of 98.5% (Fig. 7). Non-endocrinology specialists demonstrated less accuracy in detecting prediabetes and at-risk individuals than did the AI-CDSS.

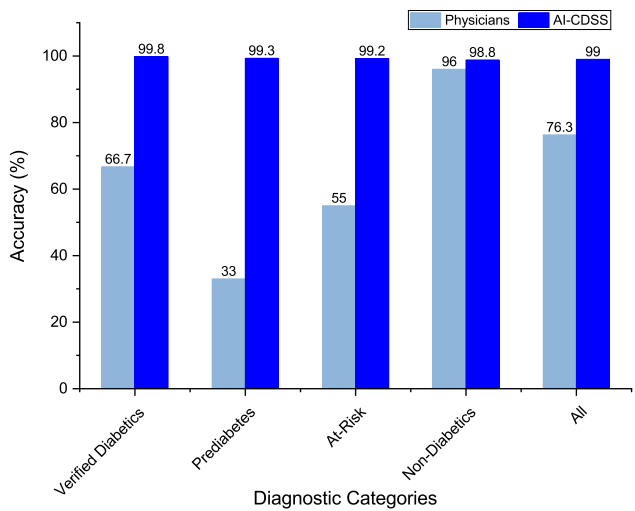

Fig. 7: A comparative study of non-endocrinology specialists (Physicians) and AI-CDSS diagnostic accuracy in a prospective cohort.

## V. DISCUSSION

The accurate diagnosis of diabetes can be challenging, even among medical experts specializing in diabetes management. To address this, we developed an AI-powered Clinical Decision Support System (AI-CDSS). Our AI-CDSS, trained on data from a diverse group of patients (including those with validated diabetes, prediabetes, at-risk individuals, and those without diabetes), demonstrated an exceptionally high diagnostic accuracy across all patient categories. This accuracy was consistently maintained in a prospective cohort undergoing diabetes screening and routine examination. In contrast, non-specialist physicians exhibited comparatively lower accuracy in diagnosing diabetes. Therefore, the AI-CDSS can be invaluable for diabetes diagnosis and management, particularly in settings with limited access to diabetes specialists. CDSS has been applied in various areas including clinical diagnosis, preventive care, and chronic disease management [22]–[24]. While offering valuable decision-making support to enhance health care quality, the CDSS has limitations and challenges. AI-CDSS generally employs ML and natural language processing techniques to extract knowledge from both structured and unstructured data [25], [26]. By combining expert-driven knowledge acquisition with machine-learning-driven rule creation, the accuracy of the system is further improved [27]. Our AI-CDSS adopts a hybrid approach to knowledge acquisition, involving three key steps: ML-driven rule development, expert-driven knowledge acquisition, and the hybridization of both. This hybrid approach has yielded significant gains in diagnostic accuracy, making it a valuable tool for early diagnosis and management of diabetes. This, in turn, could lead to improved patient outcomes and reduced strain on the healthcare system. We began by developing CKM through expert-driven knowledge acquisition, converting expert knowledge into a decision tree and mind map. This model was then compared to the PM created using machine learning methods such as random forests, decision trees, J48, CART/CRT, and CHAID, utilizing a large dataset. The transparency of these "white-box" AI models allows for easy identification of the attributes needed for classifying new patient data, simplifying the verification process, and increasing physician satisfaction by making the AI's decisions more understandable. Additionally, it reduces the computational complexity and ensures efficiency. In contrast, "black-box" models, while powerful, can be problematic in clinical settings due to their lack of transparency and interpretability. To create the final hybrid knowledge model, termed R-CKM, we combined expert-driven CKM with an ML-driven PM. This integration is essential for ensuring the accuracy and robustness of the model. The outcome of this process is the development of a web-based AI-CDSS tool for clinical use. R-CKM was converted into computer-executable and shareable machine-learning algorithms using the integrated information Authoring Tool (I-KAT), facilitating the easy application and dissemination of information in practice. Diagnosing diabetes can be challenging for medical practitioners owing to its diverse clinical presentation and overlapping symptoms with other conditions [28], [29]. Although hyperglycemia in patients with diabetes can cause symptoms such as excessive thirst, frequent urination, and unexplained weight loss, similar symptoms can also arise from other conditions such as kidney disease, infections, or even stress. Moreover, complications such as neuropathy, blindness, and cardiovascular problems can further complicate the diagnosis and management of diabetes [30], [31]. In real-world practice, patients may experience false positives for diabetes when they have other conditions and vice versa. An accurate diagnosis is critical for patient prognosis and diabetes management. Proper diagnosis and treatment can reduce healthcare costs, improve patients' quality of life, and help prevent or delay complications. Conversely, misdiagnosis can lead to ineffective treatment, an increased risk of complications, and unnecessary healthcare expenditure. The AI-CDSS has the potential to revolutionize medical decision making, reduce clinical errors, and improve the quality of life of individuals with diabetes. It can provide diagnostic support, generate alerts and reminders, evaluate and plan treatments, and assess diagnostic tests. By combining expert-driven knowledge with machine learning algorithms to enhance diagnostic accuracy and support physician decision making, the AI-CDSS can improve patient outcomes in diabetes management.

## VI. APPROACHES

This section outlines the primary data collection process, development of an AI-CDSS, and the variables used to assess the efficacy of the system in diagnosing diabetes. Fig. 8 shows the key elements involved in AI-CDSS for diabetes.

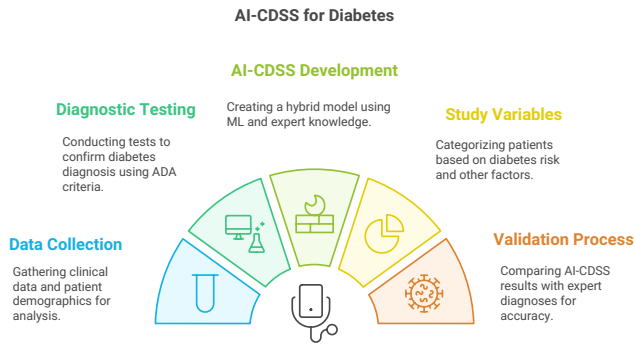

Fig. 8: The development of AI-CDSS for diabetes

### A. The Reflective Cohort

From January 2023 to December 2024, 1,298 participants with and without diabetes were included in our study. Ethical approval for data collection was obtained from Riphah International University on November 24, 2022. Two datasets containing the patient data were created. The first 650 patients served as a training dataset for ML in the creation of the AI-CDSS, whereas the subsequent 648 patients served as a test dataset for AI-CDSS validation. Patients with hyperglycemic symptoms (polyuria, polydipsia, and unexplained weight loss) or a history of diabetes mellitus were considered diabetic. Among the diagnostic standards were: 126 mg/dL (7.0 mmol/L) or above in fasting plasma glucose, an oral glucose tolerance test with a 2-hour plasma glucose level of 200 mg/dL (11.1 mmol/L) or higher, and a minimum 6.5% hemoglobin A1c level. Similarly, patients with characteristic hyperglycemia symptoms and a random plasma glucose level of ≥ 200 mg/dL (11.1 mmol/L) were considered to have diabetes. Clinical data, including demographics, signs and symptoms, and medical histories, were collected. Diagnostic tests, such as the oral glucose tolerance test and laboratory investigations, such as fasting plasma glucose and HbA1c levels, were also performed. Patients without diabetes who served as controls were randomly selected from the electronic medical records.

### B. A Potential Pilot Group

To validate the AI-CDSS, we prospectively recruited 145 consecutive patients with symptoms suggestive of diabetes. Treating physicians obtained patient histories, performed physical examinations, ordered diagnostic tests, and used clinical judgment to establish a final diagnosis (diabetes or non-diabetes). After initial entry into an electronic clinical research form, data were automatically uploaded to the AI-CDSS. Each patient in the prospective cohort provided informed consent before recruitment.

### C. Diagnostic Testing

The American Diabetes Association criteria were followed to conduct all the diabetes diagnostic tests, which involved standard medical equipment and procedures. Oral glucose tolerance testing, HbA1c levels, and fasting blood glucose levels were used to evaluate the blood sugar levels. Samples were obtained using conventional venipuncture techniques to monitor blood glucose levels, and calibrated glucometers were used to perform the measurements. To ensure accuracy and consistency, HbA1c levels were measured using high-performance liquid chromatography (HPLC). A fasting blood sample was used for the oral glucose tolerance test, which was followed by the injection of a glucose solution, and blood samples were collected regularly to track blood glucose levels over time. Furthermore, urinalysis, lipid profiles, and kidney function tests were performed to evaluate related metabolic and organ function parameters. Tests on kidney function were carried out to look for signs of diabetic nephropathy, lipid profiles were examined to track levels of cholesterol and triglycerides, and urinalysis was performed to look for any anomalies, such as proteinuria, which can point to kidney damage.

### D. Diabetes AI-CDSS Generation

The typical approach used by traditional CDSSs is expert-driven, and requires close coordination between physicians and knowledge engineers. Experts in artificial intelligence and machine learning, known as knowledge engineers, have developed fundamental ideas, examined the underlying causes of clinical problems, and efficiently conveyed domain-specific knowledge in these systems. Clinical practice standards and physician expertise are the main sources of knowledge for these systems. On the other hand, the AI-CDSS for diabetes uses patient data analyzed using ML algorithms as a crucial knowledge resource. This strategy enables support systems that are both data-driven and dynamic. Fig. 5 shows how the diabetes AI-CDSS and conventional CDSSs differ from one another.

Knowledge engineers are largely responsible for maintaining the knowledge base of the existing CDSSs. On the other hand, the AI-CDSS for diabetes takes a hybrid approach, fusing expert-driven knowledge acquisition with machine learning-driven rule creation. This integration increases the scalability and adaptability of the system by reducing its dependence on knowledge engineers. Using their knowledge and clinical guidelines, domain experts (physicians) in the AI-CDSS for diabetes have developed CKM. This model represents expert-driven information in the form of a conventional top-down decision tree. The next stage is to use different machine learning techniques to create ML-driven knowledge, or PM, based on machine learning. The construction of the R-CKM is the third stage of the process. Computer scientists use a simple, fast, iterative process called agile software development. The expert-driven CKM and ML-driven prediction models were combined at this stage. Subsequently, the accuracy and dependability of the hybrid model were confirmed using data from 650 patients in the training dataset. The ultimate result is the creation of hybrid knowledge that integrates the advantages of both the ML-driven- and expert-driven methodologies. This hybrid model is an effective tool for the diagnosis and treatment of type 2 diabetes because it is not only accurate and reliable but also more adaptable and able to change in response to new data.

### E. Study variables

The American Diabetes Association (ADA) criteria [32], which include HbA1c $\geq$ 6.5%, 2-hours plasma glucose $\geq$ 200 mg/dL during an oral glucose tolerance test (OGTT), and random plasma glucose $\geq$ 200 mg/dL in individuals with typical signs of hyperglycemia or hyperglycemic crisis, were used to define diabetes. Two endocrinologists with > ten years of professional expertise independently validated the diabetes diagnosis. Diagnoses made by these specialists were regarded as the most accurate.

Depending on their glycemic condition, the patients were divided into the following categories:

1) Verified Diabetes: Individuals who meet the ADA's definition of the disease.
2) Prediabetes: Individuals with FPG within 100 to 125 mg/dL, 2-hours plasma glucose on an OGTT between 140 and 199 mg/dL, or a HbA1c between 5.7% and 6.4%.
3) At-Risk: Individuals who do not meet the criteria for prediabetes or diabetes but have risk factors for the disease.
4) Patients without diabetes: Individuals who do not meet any of the criteria for either prediabetes or diabetes.

Additionally, the study contained a range of clinical and demographic data, including

1) Age: Broken down into age ranges (20–40, 41–60, 61–80, 81–100, etc.).
2) BMI: Both males and females are categorized as obese, overweight, normal weight, and underweight.
3) Blood pressure: Systolic and diastolic blood pressure readings are recorded.
4) Lipid profiles: These comprise triglycerides, LDL, HDL, and total cholesterol.
5) Tests for kidney function: such as estimated glomerular filtration rate (eGFR) and serum creatinine. Urinalysis was performed to check for indicators of kidney disease such as proteinuria.
6) Family history of diabetes: noted as either present or absent.
7) Lifestyle factors: such as eating habits, degree of physical activity, and smoking status
8) Comorbidities: The presence of cardiovascular illnesses, hypertension, and other relevant medical conditions.

Expert diagnosis served as the gold standard for gauging the diagnostic accuracy of AI-CDSS. Concordance was achieved when the experts and AI-CDSS arrived at an identical diagnosis (e.g., both confirmed diabetes or neither diagnosed diabetes). Any discrepancy in diagnosis was considered discordant.

## VII. CONCLUSION

In this study, we demonstrated significant improvements in diagnostic accuracy by developing and assessing an AI-CDSS for diabetes identification and diagnosis. The system excelled in classifying various diabetes stages, including confirmed diabetes, prediabetes, at-risk individuals, and individuals without diabetes. The proposed hybrid CDSS model outperformed both the conventional expert-driven and standalone ML-driven approaches. They achieved higher concordance rates and increased the diagnostic accuracy by combining ML-driven knowledge acquisition with expert-driven insights. We validated the robustness of the hybrid CDSS in real-world clinical settings. This was achieved by splitting patients into training and validation datasets in a prospective cohort analysis. Moreover, a prospective cohort of 108 consecutive patients consistently showed a high diagnostic accuracy with the hybrid CDSS, emphasizing its potential clinical application. The sensitivity and specificity measures of the hybrid approach surpassed those of both expert- and ML-driven models, indicating its effectiveness in reducing false positives and false negatives. These findings suggest that the hybrid CDSS can be a valuable tool for physicians, particularly in situations where specialized expertise is limited. The system's transparency, enabled by white-box AI models, enhances physician satisfaction by providing clear explanations of the diagnostic decisions. Additionally, the decision-tree-based algorithms of the model offer high interpretability and reduced computational complexity, making it a practical choice for healthcare applications. Overall, the hybrid CDSS represents a reliable, efficient, and scalable solution for improving patient outcomes, thereby marking substantial advancements in diabetes diagnosis. By enabling earlier and more accurate diagnoses, this system has the potential to significantly improve patient care and reduce the burden of diabetes on individuals and the health care system. Future studies should focus on expanding the capabilities of this system to encompass a broader range of metabolic and chronic conditions. Further exploration is needed to investigate how the system can integrate with other medical technologies to create a comprehensive AI-driven healthcare ecosystem.

# REFERENCE

[1] J. L. Scully, "What is a disease? Disease, disability, and their definitions." EMBO Rep., vol. 5, no. 7, pp. 650–653, 2004.
[2] N. Armstrong and P. Hilton, "Doing diagnosis: whether and how clinicians use a diagnostic tool of uncertain clinical utility," Soc. Sci. Med., vol. 120, pp. 208–214, 2014.
[3] I. Rehan, K. Rehan, S. Sultana, and M. U. Rehman, "Fingernail Diagnostics: Advancing type II diabetes detection using machine learning algorithms and laser spectroscopy," Microchem. J., p. 110762, 2024.
[4] I. Rehan, S. Khan, M. A. Gondal, Q. Abbas, and R. Ullah, "Non-invasive Diabetes Mellitus Diagnostics Using Laser-Induced Breakdown Spectroscopy and Support Vector Machine Algorithm," Arab. J. Sci. Eng., pp. 1–9, 2023.
[5] M. Abebe et al., "Variation of urine parameters among diabetic patients: A cross-sectional study," Ethiop. J. Health Sci., vol. 29, no. 1, 2019.
[6] N. Houstis, E. D. Rosen, and E. S. Lander, "Reactive oxygen species have a causal role in multiple forms of insulin resistance," Nature, vol. 440, no. 7086, pp. 944–948, 2006.
[7] P. Qian et al., "How breastfeeding behavior develops in women with gestational diabetes mellitus: A qualitative study based on health belief model in China," Front. Endocrinol. (Lausanne)., vol. 13, p. 955484, 2022.
[8] Mennickent, D., Rodríguez, A., Farías-Jofré, M., Araya, J. and Guzmán-Gutiérrez, E., 2022. Machine learning-based models for gestational diabetes mellitus prediction before 24–28 weeks of pregnancy: a review. Artificial Intelligence in Medicine 132, 102378.
[9] S. Q. Ali, S. Khalid, and S. B. Belhaouari, "A novel technique to diagnose sleep apnea in suspected patients using their ECG data," IEEE Access, vol. 7, Art. no. 8666106, pp. 35184–35194, 2019, doi: 10.1109/ACCESS.2019.2904601.
[10] I. Rehan, S. Khan, and R. Ullah, "Raman spectroscopy assisted support vector machine: a steadfast tool for noninvasive classification of urinary glucose of diabetes mellitus," Phys. Scr., vol. 99, no. 2, p. 26004, 2024.
[11] T. Ahsan, S. Khalid, S. Najam, M. A. Khan, Y. J. Kim, and B. Chang, "HRNetO: Human action recognition using unified deep features optimization framework," Comput. Mater. Contin., vol. 75, no. 1, pp. 1089–1105, 2023, doi: 10.32604/cmc.2023.034563. 10.32604/cmc.2022.025840.
[12] A. Dutta et al., "Early prediction of diabetes using an ensemble of machine learning models," Int. J. Environ. Res. Public Health, vol. 19, no. 19, p. 12378, 2022.
[13] I. Rehan, R. Ullah, and S. Khan, "Non-invasive Characterization of Glycosuria and Identification of Biomarkers in Diabetic Urine Using Fluorescence Spectroscopy and Machine Learning Algorithm," J. Fluoresc., pp. 1–9, 2023.
[14] E. A. Shah, S. J. Ahsan, M. F. Qureshi, S. Khalid, R. Ullah, and M. U. Rehman, "Machine learning and multivariate analysis of depression prevalence and predictors in acute coronary syndrome," IEEE Access, vol. 13, pp. 186231–186250, 2025, doi: 10.1109/ACCESS.2025.3624309.
[15] M. U. Rehman, M. Driss, A. Khakimov, and S. Khalid, "Non-invasive early diagnosis of obstructive lung diseases leveraging machine learning algorithms," Comput. Mater. Contin., vol. 72, no. 3, pp. 5681–5697, 2022, doi:
[16] L. Fregoso-Aparicio, J. Noguez, L. and Montesinos, J. A. García-García, "Machine learning and deep learning predictive models for type 2 diabetes: a systematic review," Diabetol. Metab. Syndr., vol. 13, no. 1, p. 148, 2021.
[17] R. Ullah, I. Rehan, and S. Khan, "Utilizing machine learning algorithms for precise discrimination of glycosuria in fluorescence spectroscopic data," Spectrochim. Acta Part A Mol. Biomol. Spectrosc., p. 124582, 2024.
[18] Eghbali-Zarch, M. and Masoud, S., 2024. Application of machine learning in affordable and accessible insulin management for type 1 and 2 diabetes: A comprehensive review. Artificial Intelligence in Medicine, p.102868.
[19] E. H. Shortliffe and M. J. Sepúlveda, "Clinical decision support in the era of artificial intelligence," Jama, vol. 320, no. 21, pp. 2199–2200, 2018.
[20] Spoladore, D., Tosi, M. and Lorenzini, E.C., 2024. Ontology-based decision support systems for diabetes nutrition therapy: A systematic literature review. Artificial Intelligence in Medicine, p.102859.
[21] R. S. Ledley and L. B. Lusted, "Reasoning foundations of medical diagnosis: symbolic logic, probability, and value theory aid our understanding of how physicians reason," Science 80, vol. 130, no. 3366, pp. 9–21, 1959.
[22] I. Kavakiotis, O. Tsave, A. Salifoglou, N. Maglaveras, I. Vlahavas, and I. Chouvarda, "Machine learning and data mining methods in diabetes research," Comput. Struct. Biotechnol. J., vol. 15, pp. 104–116, 2017.
[23] J. A. Kline, R. A. Zeitouni, J. Hernandez-Nino, and A. E. Jones, "Randomized trial of computerized quantitative pretest probability in low-risk chest pain patients: effect on safety and resource use," Ann. Emerg. Med., vol. 53, no. 6, pp. 727–735, 2009.
[24] N. Kucher et al., "Electronic alerts to prevent venous thromboembolism among hospitalized patients," N. Engl. J. Med., vol. 352, no. 10, pp. 969–977, 2005.
[25] C. L. Roumie et al., "Improving blood pressure control through provider education, provider alerts, and patient education: a cluster randomized trial," Ann. Intern. Med., vol. 145, no. 3, pp. 165–175, 2006.
[26] G. Hinton, "Deep learning—a technology with the potential to transform health care," Jama, vol. 320, no. 11, pp. 1101–1102, 2018.
[27] S. P. Somashekhar et al., "Watson for Oncology and breast cancer treatment recommendations: agreement with an expert multidisciplinary tumor board," Ann. Oncol., vol. 29, no. 2, pp. 418–423, 2018.
[28] M. Hussain et al., "Data-driven knowledge acquisition, validation, and transformation into HL7 Arden Syntax," Artif. Intell. Med., vol. 92, pp. 51–70, 2018.
[29] A. D. A. P. P. Committee and A. D. A. P. P. Committee:, "Classification and diagnosis of diabetes: Standards of Medical Care in Diabetes—2022," Diabetes Care, vol. 45, no. Supplement 1, pp. S17–S38, 2022.
[30] A. D. Deshpande, M. Harris-Hayes, and M. Schootman, "Epidemiology of diabetes and diabetes-related complications," Phys. Ther., vol. 88, no. 11, pp. 1254–1264, 2008.
[31] Spänig, S., Emberger-Klein, A., Sowa, J.P., Canbay, A., Menrad, K. and Heider, D., 2019. The virtual doctor: an interactive clinical-decision-support system based on deep learning for non-invasive prediction of diabetes. Artificial Intelligence in Medicine, 100, p.101706.
[32] D. Care, "Standards of Care in Diabetes—2023," Diabetes Care, vol. 46, pp. S1–S267, 2023.